\documentclass[journal]{IEEEtran}

\usepackage{times}
\usepackage{soul}
\usepackage{url}
\usepackage[hidelinks]{hyperref}
\usepackage[utf8]{inputenc}
\usepackage[small]{caption}
\usepackage{graphicx}
\usepackage{amsmath}
\usepackage{amsthm}
\usepackage{booktabs}
\usepackage{algorithm}
\usepackage{algorithmic}
\usepackage{multirow}
\usepackage{amsfonts}
\usepackage{fancyhdr}
\usepackage{enumitem}
\usepackage{printlen}
\usepackage{flushend}

\emergencystretch=\maxdimen
\begin{document}

\title{Adversarial Robustness Assessment: \\Why both $L_0$ and $L_\infty$ Attacks Are Necessary}

\author{Shashank~Kotyan and
        Danilo~Vasconcellos~Vargas
\thanks{S. Kotyan and D.V. Vargas are with the Laboratory of Intelligent Systems, Department
of Informatics, Kyushu University, Japan. 
\url{http://lis.inf.kyushu-u.ac.jp/}. 
E-mail:\url{vargas@inf.kyushu-u.ac.jp}}
}


\maketitle

\begin{abstract}
There exists a vast number of adversarial attacks and defences for machine learning algorithms of various types which makes assessing the robustness of algorithms a daunting task.
To make matters worse, there is an intrinsic bias in these adversarial algorithms.
Here, we organise the problems faced: 
a) Model Dependence,  b) Insufficient Evaluation, c) False Adversarial Samples, and d) Perturbation Dependent Results).
Based on this, we propose a model agnostic dual quality assessment method, together with the concept of robustness levels to tackle them.
We validate the dual quality assessment on state-of-the-art neural networks (WideResNet, ResNet, AllConv, DenseNet, NIN, LeNet and CapsNet) as well as adversarial defences for image classification problem. 
We further show that current networks and defences are vulnerable at all levels of robustness.
The proposed robustness assessment reveals that depending on the metric used (i.e., $L_0$ or $L_\infty$), the robustness may vary significantly. 
Hence, the duality should be taken into account for a correct evaluation. 
Moreover, a mathematical derivation, as well as a counter-example, suggest that $L_1$ and $L_2$ metrics alone are not sufficient to avoid spurious adversarial samples.
Interestingly, the threshold attack of the proposed assessment is a novel $L_\infty$ black-box adversarial method which requires even less perturbation than the One-Pixel Attack (only $12\%$ of One-Pixel Attack's amount of perturbation) to achieve similar results.
\end{abstract}


\IEEEpeerreviewmaketitle


\section{Introduction}
    
    \begin{figure*}[!t]
        \centering
        \textbf{ Few-Pixel ($\mathbf{L_0}$) Attack} \par\medskip
        \includegraphics[width=0.8255\textwidth]{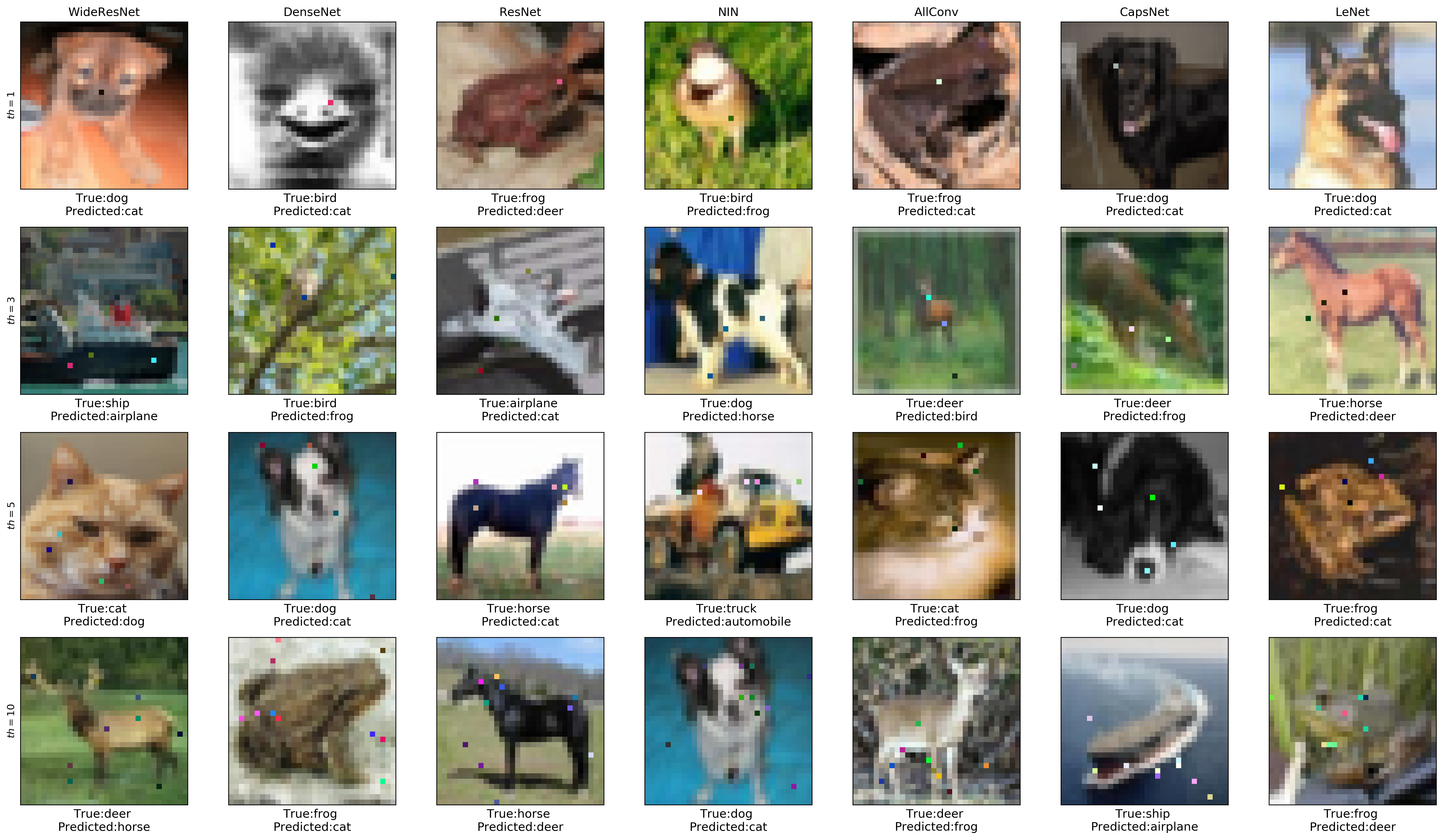} \\
        \textbf{ Threshold ($\mathbf{L_\infty}$) Attack} \par\medskip
        \includegraphics[width=0.8255\textwidth]{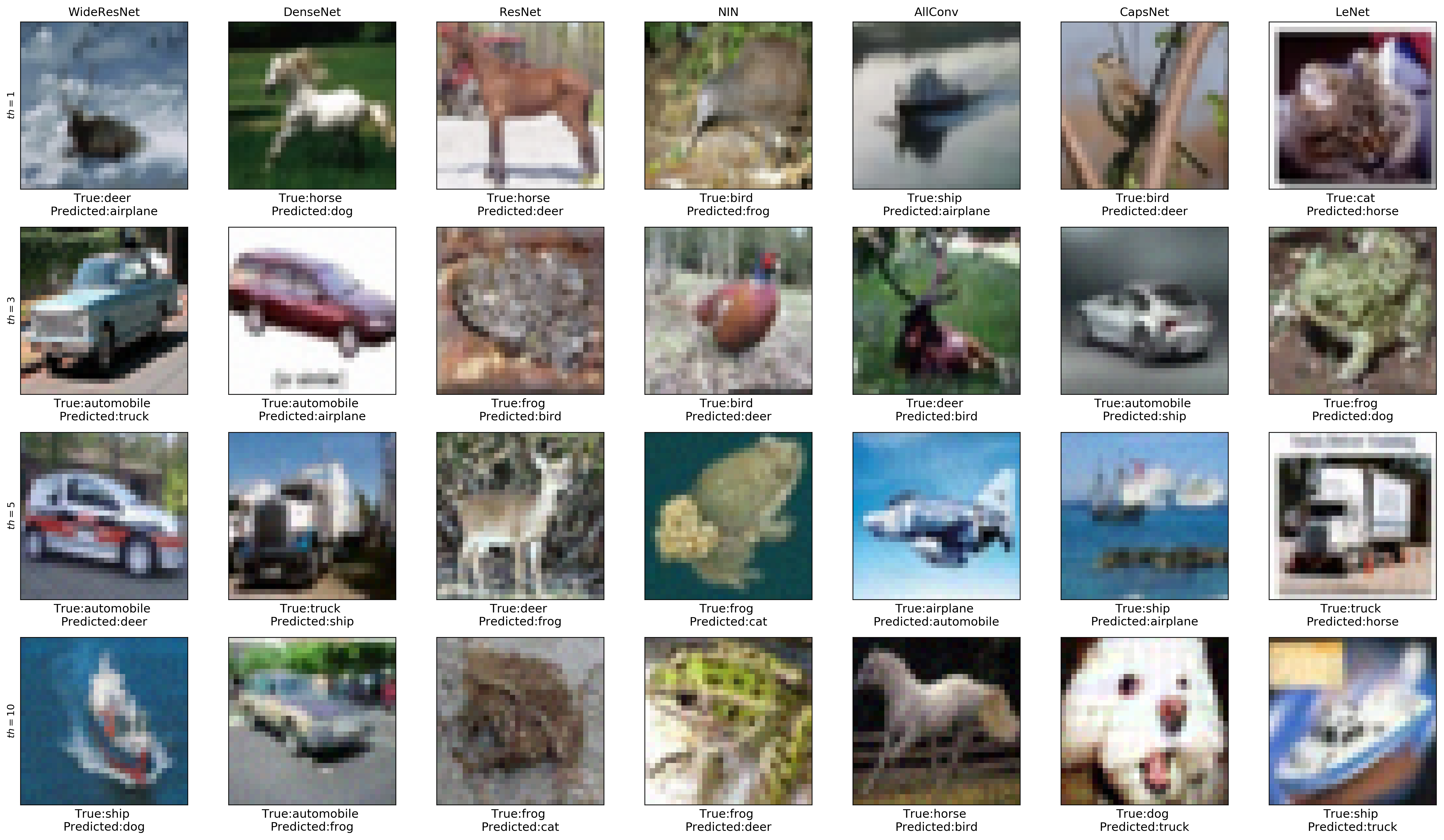}
        \caption{Adversarial samples found with Few-pixel ($L_0$) black-box attack and threshold ($L_\infty$) black-box attack.}
        \label{adv}
    \end{figure*}

    \IEEEPARstart{N}{eural} networks have empowered us to obtain high accuracy in several applications like speech recognition and face recognition.
    Most of these applications are only feasible by the aid of neural networks.
    Despite these accomplishments, neural networks have been shown to misclassify if small perturbations are added to original samples, called adversarial samples.
    Further, these adversarial samples exhibit that conventional neural network architectures are not capable of understanding concepts or high-level abstractions as we earlier speculated.
	
    Security and safety risks created by these adversarial samples is also prohibiting the use of neural networks in many critical applications such as autonomous vehicles.
    Therefore, it is of utmost significance to formulate not only accurate but robust neural networks. 
    However, to do so, a quality assessment is required, which would let robustness to be evaluated efficiently without in-depth knowledge of adversarial machine learning.
	
    Regarding the development of a quality assessment for robustness, the field of adversarial machine learning has provided some tools which could be useful for the development.
    However, the sheer amount of scenarios, adversarial attacking methods, defences and metrics ($L_0$, $L_1$, $L_2$ and $L_\infty$) make the current state-of-the-art difficult to perceive. 
    Moreover, most of the contemporary adversarial attacks are white-box ones which can not be used to assess hybrids, non-standard neural networks and other classifiers in general.
    Giving the vast amount of possibilities and many definitions with their exceptions and trade-offs, it turns out that a simple robustness quality assessment is a daunting task.
    
    Moreover, adversarial samples point out to reasoning shortcomings in machine learning.
    Improvements in robustness should also result in learning systems that can better reason over data as well as achieve a new level of abstraction.
    Therefore, a quality assessment procedure would also be helpful in this regard, checking for failures in both reasoning and high-level abstractions.
   	
    Therefore, to create a quality assessment procedure, we formalise some of the problems which must be tackled: 
    \begin{description}[itemsep=1.6pt, leftmargin=*]
    
    \item[P1 Model Dependence:] 
        A model agnostic quality assessment is crucial to enable neural networks to be compared with other approaches which may be completely different (logic hybrids and evolutionary hybrids).
    
    \item[P2 Insufficient Evaluation:] 
        There are several types of adversarial samples as well as potential attack variations and scenarios each with their own bias.
        The attacks also differ substantially depending on metrics optimized, namely $L_0$, $L_1$, $L_2$ and $L_\infty$.
        However, not all of them are vital for the evaluation of robustness.
        A quality assessment should have few but sufficient tests to provide an in-depth analysis without compromising its utility.
    
    \item[P3 False Adversarial Samples:] 
        Adversarial attacks are known sometimes to produce misleading adversarial samples (samples that can not be recognised even by a human observer) seldomly.
        Such deceptive adversarial samples can only be detected through inspection, which causes the evaluation to be error-prone.
        Both the need for inspection, together with the feasibility of fraudulent adversarial samples, should not be present.
    
    \item[P4 Perturbation Dependent Results:] 
        Varying amount of perturbation leads to varying adversarial accuracy.
        Moreover, networks differ in their sensitivity to attacks given a varied amount of perturbation.
        Consequently, this might result in double standards or hide important information.
    
    \end{description}
	
    In this article, we propose a quality assessment to tackle the problems mentioned above with the following features:
    \begin{description}[itemsep=1.6pt, leftmargin=*]

    \item[Non-gradient based Black-box Attack (Address P1):] 
        Black-box attacks are desirable for a model agnostic evaluation which does not depend on specific features of the learning process such as gradients.
        Therefore, here, the proposed quality assessment is based on black-box attacks, one of which is a novel $L_\infty$ black-box attack.
        In fact, to the knowledge of the authors, this is the first $L_\infty$ black-box Attack that does not make any assumptions over the target machine learning system.
        Figure \ref{adv} show some adversarial samples crafted with the $L_0$ and $L_\infty$ black-box Attacks used in the quality assessment.
    
    \item[Dual Evaluation (Address P2 and P3):] 
        We propose to use solely attacks based on $L_0$ and $L_\infty$ to avoid creating adversarial samples which are not correctly classified by human beings after modification. 
        These metrics impose a constraint over the spatial distribution of noise which guarantees the quality of the adversarial sample.
        In Section \ref{dual_eval}, this is explained mathematically as well as illustrated with a counter-example.
    
    \item[Robustness Levels (Address P4):] 
        In this article, we define robustness levels in terms of the constraint's threshold $th$. 
        We then compare multiple robustness levels of results with their respective values at the same robustness level.
        Robustness levels constrain the comparison of equal perturbation, avoiding the comparison of results with different degrees of perturbation (Problem P4).
        In fact, robustness levels add a concept which may aid in the classification of algorithms. 
        For example, an algorithm which is robust to One-Pixel Attack belongs to the 1-pixel-safe category.

    \end{description}

\section{Related Works}

    Recently, it was exhibited that neural networks contain many vulnerabilities.
    The first article on the topic dates back to $2013$ when it was revealed that neural networks behave oddly for almost the same images \cite{szegedy2014intriguing}.
    Afterwards, a series of vulnerabilities were found and exploited by the use of adversarial attacks. 
    In \cite{nguyen2015deep}, the authors demonstrated that neural networks show high confidence when presented with textures and random noise.
    Adversarial perturbations which can be added to most of the samples to fool a neural network was shown to be possible \cite{moosavi2017universal}. 
    Patches can also make them misclassify, and the addition of them in an image turn it into a different class \cite{brown2017adversarial}.
    Moreover, an extreme attack was shown to be effective in which it is possible to make neural networks misclassify with a single-pixel change \cite{su2019one}.
    
    Many of these attacks can be easily made into real-world threats by printing out adversarial samples, as shown in \cite{kurakin2016adversarial}.
    Moreover, carefully crafted glasses can also be made into attacks \cite{sharif2016accessorize}.
    Alternatively, even general 3D adversarial objects were shown possible \cite{athalye2017synthesizing}.
    
    Regarding understanding the phenomenon, it is argued in \cite{goodfellow2014explaining} 
    that neural networks' linearity is one of the main reasons.
    Another recent investigation proposes the conflicting saliency added by adversarial samples as the reason for misclassification \cite{vargas2019understanding}.

    Many defensive systems and detection systems have also been proposed to mitigate some of the problems.
    However, there are still no current solutions or promising ones which can negate the adversarial attacks consistently.
    Regarding defensive systems, defensive distillation in which a smaller neural network squeezes the content learned by the original one was proposed as a defence \cite{papernot2016distillation}. 
    However, it was shown not to be robust enough in \cite{carlini2017towards}.
    Adversarial training was also proposed, in which adversarial samples are used to augment the training dataset \cite{goodfellow2014explaining,huang2015learning,madry2017towards}. 
    Augmentation of the dataset is done in such a way that the neural network should be able to classify the adversarial samples, increasing its robustness.  
    Although adversarial training can increase the robustness slightly, the resulting neural network is still vulnerable to attacks \cite{tramer2017ensemble}.
    There are many recent variations of defenses \cite{dziugaite2016study,hazan2016perturbations,das2017keeping,guo2017countering,song2017pixeldefend,xu2017feature,ma2018characterizing,buckman2018thermometer} which are carefully analysed and many of their shortcomings are explained in \cite{athalye2018obfuscated,uesato2018adversarial}.
    
    Regarding detection systems, a study from \cite{grosse2017statistical} demonstrated that indeed some adversarial samples have different statistical properties which could be exploited for detection.
    In \cite{xu2017feature}, the authors proposed to compare the prediction of a classifier with the prediction of the same input but "squeezed".
    This technique allowed classifiers to detect adversarial samples with small perturbations.
    Many detection systems fail when adversarial samples deviate from test conditions \cite{carlini2017adversarial,carlini2017magnet}.
    Thus, the clear benefits of detection systems remain inconclusive.

\section{Adversarial Machine Learning As Optimisation Problem}

    Adversarial machine learning can be perceived as a constrained optimisation problem.
    Before defining it, let us formalise adversarial samples first.
    Let $f(x) \in [\![0, 1]\!]$ be the output of a machine learning algorithm in binary classification setting.
    Extrapolating the algorithm in multi-label classification setting, the output can be defined as $f(x) \in [\![1..N]\!]$.
    Here, $x \in \mathbb{R}^{k}$ is the input of the algorithm for the input of size $k$ and $N$ is the number of classes in which $x$ can be classified.
    An adversarial sample $x'$ for an original sample $x$ can be thus, defined as follows:
    \begin{equation*}  \begin{aligned}
    x' = x + \epsilon_{x} \quad \text{such that} \quad f(x') \ne f(x)
    \end{aligned} \end{equation*} 
    in which $\epsilon_{x} \in \mathbb{R}^{k}$ is a small perturbation added to the input.
    Therefore, adversarial machine learning can be defined as an optimization problem\footnote{Here the definition will only concern untargeted attacks in classification setting but a similar optimization problem can be defined for targeted attacks}:
    \begin{equation*} \begin{aligned}
    \underset{\epsilon_{x}}{\text{minimize}} \quad g(x+\epsilon_{x})_c \quad \text{subject to} \quad \Vert \epsilon_{x} \Vert \leq th
    \label{adv_eqn}
    \end{aligned} \end{equation*}
    where $th$ is a pre-defined threshold value and $g()_c$ is the soft-label or the confidence for the correct class $c$ such that $f(x) = \text{argmax } g(x)$.
    
    The constraint in the optimisation problem has the objective of disallowing perturbations which could make $x$ unrecognisable or change its correct class.
    Therefore, the constraint is itself a mathematical definition of what constitutes an imperceptible perturbation.
    Many different norms are used in the literature (e.g., $L_0$, $L_1$, $L_2$ and $L_\infty$).
    Intuitively, the norms allow for different types of attacks.

    For simplicity, we are narrowing the scope of this article to the image classification problem alone.
    However, the proposed attacks and the quality assessment can be also be extended to other problems as well.
    
\section{Guaranteeing the Quality of Adversarial Samples}
\label{dual_eval}

    \begin{figure}[!t]
    \centering
    \includegraphics[width=0.40\textwidth]{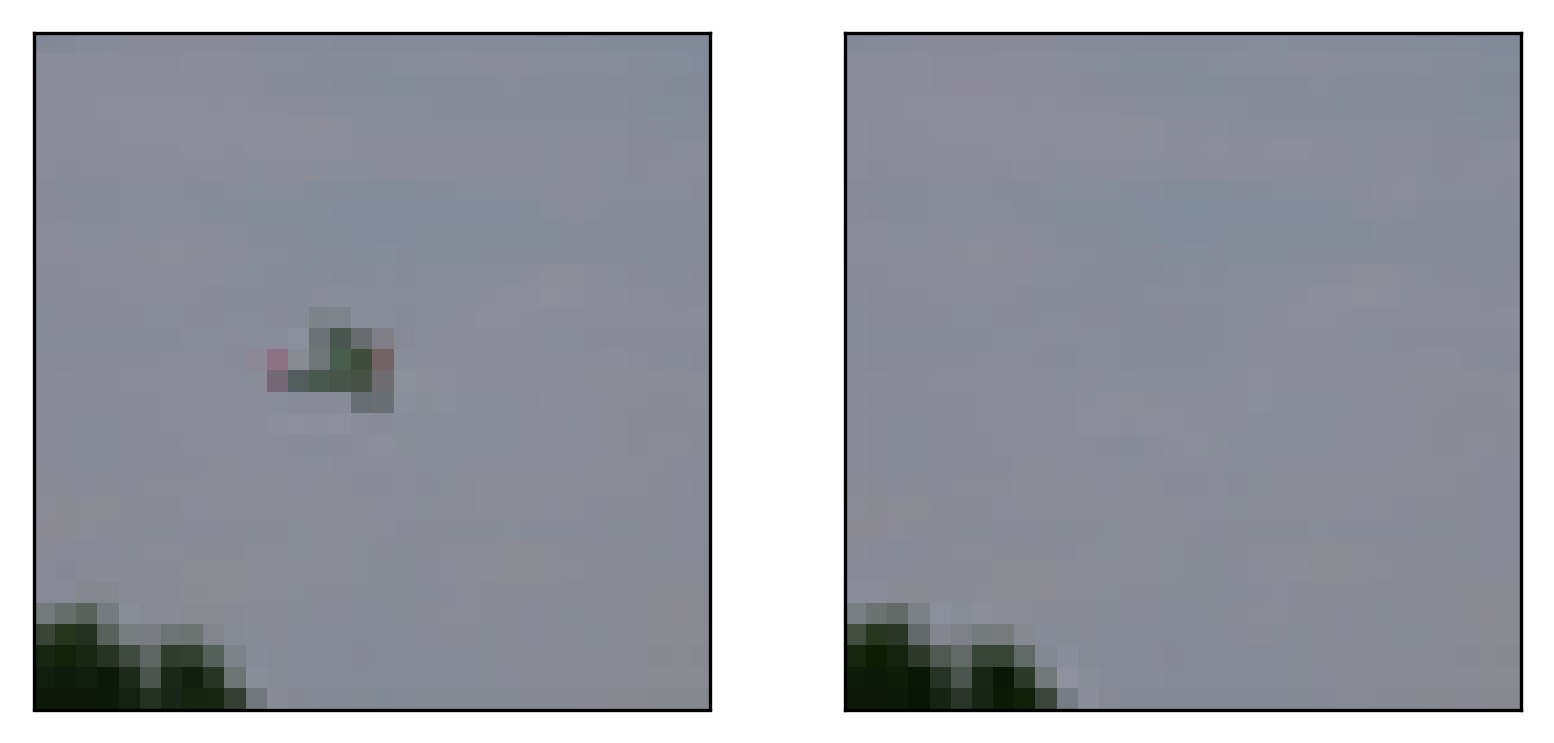} 
    \caption{
        Example of a false adversarial sample (right) and its respective original sample (left). 
        The false adversarial sample is built with few total perturbations (i.e., low $L_1$ and $L_2$) but with unrecognisable final image (false adversarial sample). 
        This is a result of the non-constrained spatial distribution of perturbations which is prevented if low $L_0$ or $L_\infty$ is used.
        This hypothetical attack has a $L_2$ of merely $356$, well below the maximum $L_2$ for the One-Pixel ($L_0 \leq 1$) Attack ($765$).
    }
    \label{example}
    \end{figure}

    Constraining the perturbation is decisive in adversarial samples to avoid producing samples that can not be recognised by human beings or samples that have, by the amount of perturbation, changed its correct class.
    However, restraining the total amount of perturbation is not enough as a small amount of perturbation concentrated in a few pixels might be able to create false adversarial samples.
    Therefore, a spatial constraint over the perturbation of pixels $P$ would be a desirable feature.
    
    This can be achieved mathematically as follows:
    Given an image $x$ and its perturbed counterpart $x' $, it is possible to calculate $L_1$ norm between the original by the Manhattan distance of both matrices:
    $ \Vert x - x' \Vert_1 $.
    Constraining $L_1$ to be less than a certain number does not guarantee any spatial distribution constraint.
    Let us define a set based on all non-zero pixel perturbations as follows:
    $N_z = \{ P_i : \Vert P_i - P_i' \Vert_1 > 0 \}$
    where $P_i$ and $P_i'$ are pixels from respectively the original image $x$ and the perturbed image $x'$ and $i$ is an image index.
    Both $N_z$ and its cardinality $ | N_z | $ has information about the spatial distribution of perturbations and constraining any of these values would result in a spatially limited perturbation.

    Provided that $th$ is low enough, a modification preserving the white noise of that intensity would bound $ | N_z | < th$.
    Moreover, $ | N_z | $ is precise $L_0$, demonstrating that $L_0$ is based on the set $N_z$, which stores spatial information about the differences. 
    At the same time, $L_1$ uses the Manhattan norm, which does not have this information.
    Similarly, the $L_\infty$ norm can be rewritten as the following optimisation constraint: 
    $\forall P_i \in x, \Vert P_i - P_i' \Vert_\infty \leq th $
    Notice that this constraint is also defined over the spatial distribution of perturbations.
    
    Figure \ref{example} gives empirical evidence of misleading adversarial sample of an image that is constrained by $L_2 \leq 765$.
    Notice that this value is precisely the maximum change of one pixel, i.e., the maximum possible perturbation of the One-Pixel attack ($L_0 \leq 1$) which when no limits are imposed over its spatial distribution may create false adversarial samples.

    The reasoning behind the $L_0$ and $L_\infty$ are as follows, without altering much the original sample, attacks can perturb a few pixels strongly ($L_0$), all pixels slightly ($L_\infty$) or a mix of both ($L_1$ and $L_2$).
    The hurdle is that $L_1$ and $L_2$ which mix both strategies vary strongly with the size of images, if not used with caution may cause unrecognisable adversarial samples (Problem P3).
    Also, it is difficult to compare between methods using $L_1$ and $L_2$ norm because the amount of perturbations will often differ (Problem P4). 
    
    \begin{description}[itemsep=1.6pt, leftmargin=*]
    
    \item[Threshold Attack ($L_\infty$ black-box Attack):]
        The threshold attack optimizes the constrained optimization problem with the constraint $\Vert \epsilon_{x} \Vert_\infty \leq th$, i.e., it uses the $L_\infty$ norm.
        The algorithm search in $\mathbb{R}^{k}$ space as the search space is the same as the input space.
        This is because the variables can be any variation of the input as long as the threshold is respected.
        In image classification problem $k = m \times n \times c$ where $m \times n$ is the size, and $c$ is the number of channels of the image.
        
    \item[Few-Pixel Attack ($L_0$ black-box Attack):]
        The few-pixel attack is a variation of our previous proposed attack, the One-Pixel Attack \cite{su2019one}.
        It optimizes the constrained optimization problem by using the constraint $\Vert \epsilon_{x} \Vert_0 \leq th$, i.e., it uses the $L_0$ norm.
        The search variable is a combination of pixel values (depending on channels $c$ in the image) and position ($2$ values X, Y) for all of the pixels ($th$ pixels).
        Therefore, the search space is smaller than the threshold attack defined below with dimensions of $\mathbb{R}^{(2+ c ) \times th}$.

    \item[Robustness Levels:]
        Here we propose robustness levels, as machine learning algorithms might perform differently to varying amount of perturbations.
        Robustness levels evaluate classifiers in a couple of $th$ thresholds.
        Explicitly, we define four levels of robustness ${1,3,5,10}$ for both of our $L_0$ Norm Attack and $L_\infty$ Norm Attack.
        We then name them respectively pixel and threshold robustness levels.
        Algorithms that pass a level of robustness ($0\%$ adversarial accuracy) are called level-threshold-safe or level-pixel-safe.
        For example, an algorithm that passes the level-one in threshold ($L_\infty$) attack is called 1-threshold-safe.

    \end{description}

\section{Experimental Results and Discussions}

    In this section, we aim to validate the dual quality assessment\footnote{Code is available at \url{http://bit.ly/DualQualityAssessment}} empirically as well as analyse the current state-of-the-art neural networks in terms of robustness.

    \begin{description}[itemsep=1.6pt, style=sameline, leftmargin=*]
        
    \item[Preliminary Tests (Section \ref{section_preliminary_tests}):]  
        Tests on two state-of-the-art neural networks are presented (ResNet \cite{he2016deep} and CapsNet \cite{sabour2017dynamic}). 
        These tests are done to choose the black-box optimisation algorithm to be used for the further sections. 
        The performance of both Differential Evolution (DE) \cite{storn1997differential} and Covariance Matrix Adaptation Evolution Strategy (CMA-ES) \cite{hansen2003reducing} are evaluated.
    
    \item[Evaluating Learning and Defense Systems (Section \ref{section_evaluating_systems}):]  
        Tests are extended to the seven different state-of-the-art neural networks - 
        WideResNet \cite{zagoruyko2016wide}, 
        DenseNet \cite{iandola2014densenet}, 
        ResNet \cite{he2016deep}, 
        Network in Network (NIN) \cite{lin2013network}, 
        All Convolutional Network (AllConv) \cite{springenberg2014striving}, 
        CapsNet \cite{sabour2017dynamic}, and 
        LeNet \cite{lecun1998gradient}.
        We also evaluate three adversarial defences applied to the standard ResNet architecture - 
        Adversarial training (AT) \cite{madry2017towards}, 
        Total Variance Minimization (TVM) \cite{guo2017countering}, and 
        Feature Squeezing (FS) \cite{xu2017feature}.
        We have chosen defences based on entirely different principles to be tested.
        In this way, the results achieved here can be extended to other similar types of defences in the literature.
    
    \item[Evaluating Other Adversarial Attacks (Section \ref{section_comparison}):]
        The evaluated learning systems are tested against other existing white-box and black-box adversarial attacks such as - 
        Fast Gradient Method (FGM) \cite{goodfellow2014explaining}, 
        Basic Iterative Method (BIM) \cite{kurakin2016adversarial}, 
        Projected Gradient Descent Method (PGD) \cite{madry2017towards}, 
        DeepFool \cite{moosavi2016deepfool}, and 
        NewtonFool \cite{jang2017objective}. 
        This analysis further helps to demonstrate the necessity of duality in quality assessment.
        
    \item[Extremely Fast Quality Assessment (Section \ref{section_transferability}):] 
        In this section, we apply and evaluate the principle of transferability of adversarial samples.
        We verify the possibility of a speedy version of the proposed quality assessment. 
        We implement this by using already crafted adversarial samples to fool neural networks, instead of a full-fledged optimisation.
        This would enable attacks to have a $O(1)$ time complexity, being significantly faster.
    
    \item[Quality Assessment's Attack Distribution (Section \ref{section_distribution}):]
        Here, we assess the dual-attack distribution (Few-Pixel Attack and Threshold Attack).
        The analysis of the distribution demonstrates the necessity of such duality.
        The distribution of successful attacks are shown, and previous attacks are analysed in this perspective. 
    
    \item[Effect of threshold (Section \ref{section_analysing_systems}):]
        We analyse the complete behaviour of the adversarial accuracy of our black-box attacks without restricting the threshold's $th$ value.
        Using this analysis, we prove the results using a fixed $th$ in robustness levels is a reasonable approximation for our proposed quality assessment.

    \end{description}

    \subsection{Experimental Settings}

        \begin{table}[!t]
            \centering
            \resizebox{\columnwidth}{!}{
            \begin{tabular}{ll|l}
            \toprule
            \multicolumn{2}{l|}{\textbf{Attack}} & \textbf{Parameters} \\
            \midrule
            \multicolumn{2}{l|}{FGM}                                & $\text{norm} = L_\infty$, $\epsilon=8$, $\epsilon_{\text{step}}=2$  \\  
            \multicolumn{2}{l|}{BIM}                                & $\text{norm} = L_\infty$, $\epsilon=8$, $\epsilon_{\text{step}}=2$, $\text{iterations} = 10$ \\ 
            \multicolumn{2}{l|}{PGD}                                & $\text{norm} = L_\infty$, $\epsilon=8$, $\epsilon_{\text{step}}=2$, $\text{iterations} = 20$ \\ 
            \multicolumn{2}{l|}{DeepFool}                           & $\text{iterations} = 100$, $\epsilon=0.000001$                      \\       
            \multicolumn{2}{l|}{NewtonFool}                         & $\text{iterations} = 100$, $\text{eta}=0.01$                        \\    
            \midrule
            \multirow{3}{*}{$L_0$ Attack}      & Common & $\text{Parameter Size} = 5$, \\
                                                           & DE     & $\text{NP} = 400$, $\text{Number of Generations} = 100$,  $\text{CR} = 1$\\
                                                           & CMA-ES & $\text{Function Evaluations} = 40000$, $\sigma = 31.75$ \\
            
            \midrule 
            \multirow{3}{*}{$L_\infty$ Attack} & Common & $\text{Parameter Size} = 3072$, \\
                                                           & DE     & $\text{NP} = 3072$, $\text{Number of Generations} = 100$,  $\text{CR} = 1$\\
                                                           & CMA-ES & $\text{Function Evaluations} = 39200$, $\sigma = th/4$ \\
            \bottomrule
            \end{tabular}
            }
            \caption{Description of various parameters of different adversarial attacks.}
            \label{parameter}
        \end{table}

        We use CIFAR-10 dataset \cite{krizhevsky2009learning} to evaluate our dual quality assessment.
        Table \ref{parameter} gives the parameter description of various adversarial attacks used.
        All the pre-existing adversarial attacks used in the article have been evaluated using Adversarial Robustness 360 Toolbox (ART v1.2.0) \cite{art2018}.  

        For our $L_0$ and $L_\infty$ Attacks, we use the canonical versions of the DE and CMA-ES algorithms to have a clear standard.
        DE uses a repair method in which values that go beyond range are set to random points within the valid range.
        While in CMA-ES, to satisfy the constraints, a simple repair method is employed in which pixels that surpass the minimum/maximum are brought back to the minimum/maximum value.
        Moreover, a clipping function is used to keep values inside the feasible region.
        The constraint is always satisfied because the number of parameters is itself modelled after the constraint.
        In other words, when searching for one pixel perturbation, the number of variables are fixed to pixel values (three values) plus position values (two values).
        Therefore it will always modify only one pixel, respecting the constraint.
        Since the optimisation is done in real values, to force the values to be within range, a simple clipping function is used for pixel values.
        For position values, a modulo operation is executed.
        
    \subsection{Preliminary Tests: Choosing the Optimization Algorithm}
    \label{section_preliminary_tests}

        \begin{table}[!t]
            \centering
            \resizebox{\columnwidth}{!}{%
            \begin{tabular}{ll|rrrr}
            \toprule
            \multirow{2}{*}{\textbf{Model}} & \multicolumn{1}{c|}{\textbf{Attack}} & \multicolumn{4}{c}{\textbf{Adversarial Accuracy}} \\
            & \multicolumn{1}{c|}{\textbf{Optimiser}} & $\mathbf{th=1}$ & $\mathbf{th=3}$ & $\mathbf{th=5}$ & $\mathbf{th=10}$ \\
            \midrule
            \midrule
            \multicolumn{6}{c}{\textbf{Few-Pixel ($\mathbf{L_0}$) Attack}} \\
            \midrule
            \multirow{2}{*}{ResNet}     & DE     & \textbf{24\%}   & \textbf{70\%} & \textbf{75\%} & 79\%          \\ 
                                        & CMA-ES & 12\%            & 52\%          & 73\%          & \textbf{85\%} \\ 
            \multirow{2}{*}{CapsNet}    & DE     & \textbf{21\%}   & 37\%          & \textbf{49\%} & \textbf{57\%} \\ 
                                        & CMA-ES & 20\%            & \textbf{39\%} & 40\%          & 41\%          \\
            \midrule 
            \midrule
            \multicolumn{6}{c}{\textbf{Threshold ($\mathbf{L_\infty}$) Attack}} \\ 
            \midrule
            \multirow{2}{*}{ResNet}     & DE     & 5\%             & 23\%          & 53\%          & 82\%          \\
                                        & CMA-ES & \textbf{33\%}   & \textbf{71\%} & \textbf{76\%} & \textbf{83\%} \\
            \multirow{2}{*}{CapsNet}    & DE     & 11\%            & 13\%          & 15\%          & 23\%          \\
                                        & CMA-ES & \textbf{13\%}   & \textbf{34\%} & \textbf{72\%} & \textbf{97\%} \\
            \bottomrule
            \end{tabular}
            }
            \caption{
            Adversarial accuracy results for Few-Pixel ($L_0$) and Threshold ($L_\infty$) Attacks with DE and CMA-ES}
            \label{table_attack_prelim}
        \end{table}

        Table \ref{table_attack_prelim} shows the adversarial accuracy results performed over $100$ random samples.
        Here adversarial accuracy corresponds to the accuracy of the adversarial attack to create adversarial samples to fool neural networks.
        Both black-box attacks can craft adversarial samples in all levels of robustness.
        This fact demonstrates that without knowing anything about the learning system and in a constrained setting, black-box attacks are still able to reach more than $80\%$ adversarial accuracy in state-of-the-art neural networks. 
        
        Concerning the comparison of CMA-ES and DE, the outcomes favour the choice of CMA-ES for the quality assessment.
        Both CMA-ES and DE perform likewise for the Few-Fixel Attack, with both DE and CMA-ES having the same number of wins.
        However, for the Threshold Attack, the performance varies significantly.
        CMA-ES this time always wins (eight wins) against DE (no win).
        This domination of CMA-ES is expected since the Threshold Attack has a high dimensional search space which is more suitable for CMA-ES.
        This happens in part because DE's operators may allow some variables to converge prematurely.
        CMA-ES, on the other hand, is always generating slightly different solutions while evolving a distribution.

        In these preliminary tests, CapsNet was shown overall superior to ResNet.
        Few-pixel ($L_0$) Attack reach $85\%$ adversarial accuracy for ResNet when ten pixels are modified.
        CapsNet, on the other hand, is more robust to Few-Pixel Attacks, allowing them to reach only $52\%$ and $41\%$ adversarial accuracy when ten pixels are modified for DE and CMA-ES respectively.
        CapsNet is less robust than ResNet to the Threshold Attack with $th=10$ in which almost all images were vulnerable ($97\%$).
        At the same time, CapsNet is reasonably robust to 1-threshold-safe (only $13\%$ adversarial accuracy).
        ResNet is almost equally not robust throughout, with low robustness even when $th=3$, losing to CapsNet in robustness in all other values of $th$ of the threshold attack.
        These preliminary tests also show that different networks have different robustness.
        This is not only regarding the type of attacks ($L_0$ and $L_\infty$) but also with the degree of attack (e.g., 1-threshold and 10-threshold attacks have very different results on CapsNet).

    \subsection{Evaluating Learning and Defense Systems}
    \label{section_evaluating_systems}

        \begin{table}[!t]
        \centering
        \resizebox{\columnwidth}{!}{%
        \begin{tabular}{lr|rrrr}
            \toprule
            \multicolumn{2}{c|}{\textbf{Model and}} & \multicolumn{4}{c}{\textbf{Adversarial Accuracy}} \\
            \multicolumn{2}{c|}{\textbf{Standard Accuracy}} & $\mathbf{th=1}$ & $\mathbf{th=3}$ & $\mathbf{th=5}$ & $\mathbf{th=10}$ \\
            \midrule
            \midrule
            \multicolumn{6}{c}{\textbf{Few-Pixel ($\mathbf{L_0}$) Attack}} \\   
            \midrule                  
            WideResNet & 95.12\% & \textbf{11\%}  & 55\%          & 75\%          & 94\%          \\
            DenseNet   & 94.54\% & \textbf{9\%}   & 43\%          & 66\%          & 78\%          \\
            ResNet     & 92.67\% & \textbf{12\%}  & 52\%          & 73\%          & 85\%          \\
            NIN        & 90.87\% & 18\%           & 62\%          & 81\%          & 90\%          \\
            AllConv    & 88.46\% & \textbf{11\%}  & \textbf{31\%} & 57\%          & 77\%          \\
            CapsNet    & 79.03\% & 21\%           & 37\%          & \textbf{49\%} & \textbf{57\%} \\
            LeNet      & 73.57\% & 58\%           & 86\%          & 94\%          & 99\%          \\
            \midrule
            AT         & 87.11\% & 22\% & 52\% & 66\% & 86\% \\
            TVM        & 47.55\% & 16\% & 12\% & 20\% & 24\% \\
            FS         & 92.37\% & 17\% & 49\% & 69\% & 78\% \\
            \midrule
            \midrule
            \multicolumn{6}{c}{\textbf{Threshold ($\mathbf{L_\infty}$) Attack}} \\
            \midrule
            WideResNet & 95.12\% & 15\%           & 97\%          & 98\%          & 100\%         \\
            DenseNet   & 94.54\% & 23\%           & 68\%          & \textbf{72\%} & \textbf{74\%} \\
            ResNet     & 92.67\% & 33\%           & 71\%          & \textbf{76\%} & 83\%          \\
            NIN        & 90.87\% & \textbf{11\%}  & 86\%          & 88\%          & 92\%          \\
            AllConv    & 88.46\% & \textbf{9\%}   & 70\%          & \textbf{73\%} & \textbf{75\%} \\
            CapsNet    & 79.03\% & \textbf{13\%}  & \textbf{34\%} & \textbf{72\%} & 97\%          \\
            LeNet      & 73.57\% & 44\%           & 96\%          & 100\%         & 100\%         \\
            \midrule
            AT         & 87.11\% & 3\%  & 12\% & 25\% & 57\% \\
            TVM        & 47.55\% & 4\%  & 4\%  & 6\%  & 14\% \\
            FS         & 92.37\% & 26\% & 63\% & 66\% & 74\% \\
            \bottomrule
        \end{tabular}
        }
        \caption{
        Adversarial accuracy results for $L_0$ and $L_\infty$ Attacks over $100$ random samples}
        \label{table_attack}
        \end{table}

        Table \ref{table_attack} extends the CMA-ES attacks on various neural networks: 
        WideResNet \cite{zagoruyko2016wide}, 
        DenseNet \cite{iandola2014densenet}, 
        ResNet \cite{he2016deep}, 
        Network in Network (NIN) \cite{lin2013network}, 
        All Convolutional Network (AllConv) \cite{springenberg2014striving}, 
        CapsNet \cite{sabour2017dynamic}, and 
        LeNet \cite{lecun1998gradient}.
        We also evaluate with three contemporary defences: 
        Adversarial training (AT) \cite{madry2017towards}, 
        Total Variance Minimization (TVM) \cite{guo2017countering}, and 
        Feature Squeezing (FS) \cite{xu2017feature}.

        Results in bold (Only for learning systems and not defensive systems) are the lowest adversarial accuracy and other results which are within a distance of five from the lowest one. 
        For CapsNet only $88$ samples could be attacked with maximum $th=127$ for $L_0$ Attack. 
        Twelve samples could not be overwhelmed when the $th<128$.
        Here, taking into account an existing variance of results, we consider results within five of the lowest to be equally good.
        If we consider the number of bold results for each of the neural networks, a qualitative measure of robustness CapsNet and AllConv can be considered the most robust with five bold results.
        The third place in robustness achieves only three bold results and consequently is far away from the prime performers. 
        
        Regarding the adversarial training, it is easier to attack with the Few-Pixel Attack than with Threshold Attack.
        This result should derive from the fact that the adversarial samples used in adversarial training contained images from Projected Gradient Descent (PGD) Attack, which is $L_\infty$ type of attack.
        Therefore, it suggests that \textit{given an attack bias that differs from the invariance bias used to train the networks, the attack can easily succeed}.
        Regarding TVM, the attacks were less successful. 
        We trained a ResNet on TVM modified images and, albeit many trials with different hyper-parameters, we were able to craft a classifier with at best $47.55\%$ accuracy.
        This is a steep drop from the $92.37\%$ accuracy of the original ResNet and happens because TVM was initially conceived for Imagenet and did not scale well to CIFAR-10.
        However, as the original accuracy of the model trained with TVM is also not high; therefore, even with a small attack percentage of $24\%$, the resulting model accuracy is $35\%$.
        Attacks on Feature Squeezing had relatively high adversarial accuracy both $L_0$ and $L_\infty$ attacks.
        Moreover, both types of attacks had similar accuracy, revealing a lack of bias in the defence system.
    
        Notice that none of the neural networks was able to reduce low $th$ attacks to zero.
        This illustrates that although robustness may differ between current neural networks, none of them can effectively overcome even the lowest level of perturbation feasible.
        Moreover, since a $th=5$ is enough to achieve around $70\%$ accuracy in many settings, this suggests that achieving $100\%$ adversarial accuracy may depend more on a few samples which are harder to attack, such as samples far away from the decision boundary.
        Consequently, the focus on $100\%$ adversarial accuracy rather than the amount of threshold might give preference to methods which set a couple of input projections far away from others without improving the accuracy overall.
        An example can be examined by making some input projections far away enough to make them harder to attack. 

        The difference in the behaviour of $L_0$ and $L_\infty$ Norm Attacks shows that the robustness is achieved with some trade-offs.
        This further justifies the importance of using both metrics to evaluate neural networks.

    \subsection{Evaluating Other Adversarial Attacks}
    \label{section_comparison}

        \begin{table*}[!t]
        \centering
        \resizebox{2\columnwidth}{!}{%
        \begin{tabular}{ll|rrrrrrr}
            \toprule
            \multicolumn{2}{c}{\textbf{Adversarial Attacks}} & \multicolumn{1}{|c}{\textbf{WideResNet}} & \multicolumn{1}{c}{\textbf{DenseNet}} & \multicolumn{1}{c}{\textbf{ResNet}} & \multicolumn{1}{c}{\textbf{NIN}}
             & \multicolumn{1}{c}{\textbf{AllConv}} & \multicolumn{1}{c}{\textbf{CapsNet}} & \multicolumn{1}{c}{\textbf{LeNet}} \\
            \midrule
            FGM                                                       & & 69\% (159.88) & 50\% (120.03) & 52\% (124.70) & 72\% (140.46) & 67\% (155.95) & 70\% ~(208.89) & 84\% (152.37) \\
            BIM                                                       & & 89\% (208.44) & 52\% (160.34) & 55\% (164.64) & 74\% (216.97) & 69\% (273.90) & 82\% ~(361.63) & 89\% (345.27) \\
            PGD                                                       & & 89\% (208.49) & 52\% (160.38) & 55\% (164.64) & 74\% (216.96) & 69\% (274.15) & 84\% ~(370.90) & 89\% (357.34) \\
            DeepFool                                                  & & 60\% (613.14) & 60\% (478.03) & 58\% (458.57) & 59\% (492.90) & 51\% (487.46) & 87\% ~(258.08) & 31\% (132.32) \\ 
            NewtonFool                                                & & 82\% ~(63.13) & 50\% ~(53.89) & 54\% ~(51.56) & 66\% ~(54.78) & 61\% ~(61.05) & 90\% (1680.83) & 84\% ~(49.61) \\
            \midrule  
            \multirow{4}{*}{Few-Pixel ($L_0$) Attack}         & $th=1$  & 20\% (181.43) & 20\% (179.48) & 29\% (191.73) & 28\% (185.09) & 24\% (172.01) & 29\% ~(177.86) & 61\% (191.69) \\
                                                              & $th=3$  & 54\% (276.47) & 50\% (270.50) & 63\% (275.57) & 62\% (274.91) & 49\% (262.66) & 43\% ~(247.97) & 89\% (248.21) \\
                                                              & $th=5$  & 75\% (326.14) & 68\% (315.53) & 79\% (314.27) & 81\% (318.71) & 67\% (318.99) & 52\% ~(300.19) & 96\% (265.18) \\
                                                              & $th=10$ & 91\% (366.60) & 81\% (354.42) & 90\% (342.56) & 93\% (354.61) & 81\% (365.10) & 63\% ~(359.55) & 98\% (271.90) \\
            \midrule 
            \multirow{4}{*}{Threshold ($L_\infty$) Attack}    & $th=1$  & 30\% ~(39.24) & 38\% ~(39.24) & 43\% ~(39.27) & 23\% ~(39.23) & 23\% ~(39.21) & 13\% ~(39.09) & 47\%  ~(39.28) \\
                                                              & $th=3$  & 92\% ~(65.07) & 69\% ~(53.89) & 74\% ~(52.82) & 81\% ~(72.29) & 72\% ~(68.11) & 34\% ~(70.79) & 96\%  ~(62.86) \\
                                                              & $th=5$  & 95\% ~(67.84) & 72\% ~(56.81) & 77\% ~(55.38) & 85\% ~(77.09) & 76\% ~(72.45) & 72\% (130.80) & 99\%  ~(66.42) \\
                                                              & $th=10$ & 98\% ~(70.70) & 78\% ~(67.63) & 83\% ~(64.50) & 90\% ~(84.20) & 79\% ~(77.76) & 97\% (184.93) & 100\% ~(66.65) \\
            \bottomrule
        \end{tabular}
            }
        \caption{
        Adversarial accuracy of the proposed $L_0$ and $L_\infty$ black-box Attacks used in the dual quality assessment and their comparison with other methods from the literature. 
        The value in the brackets represents the Mean $L_2$ score of the adversarial sample with the original sample.
        \textit{The results were drawn by attacking a different set of samples from previous tests. 
        Therefore the accuracy results may differ slightly from previous tables.}
        }
        \label{few_table}
        \end{table*}
    
    	\begin{figure*}[!t]
    		\centering
    		\includegraphics[width=0.45\textwidth]{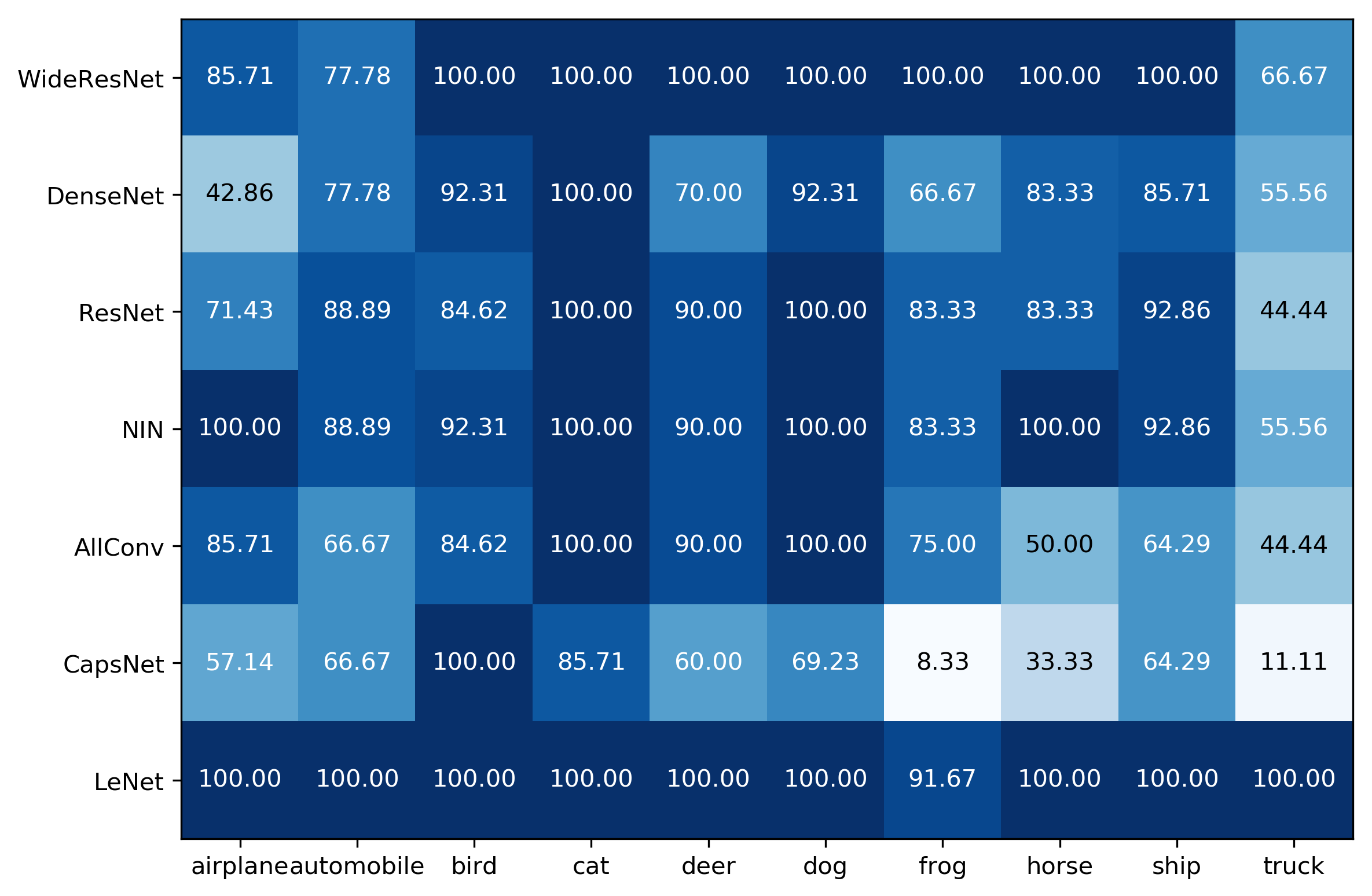}
    		\includegraphics[width=0.45\textwidth]{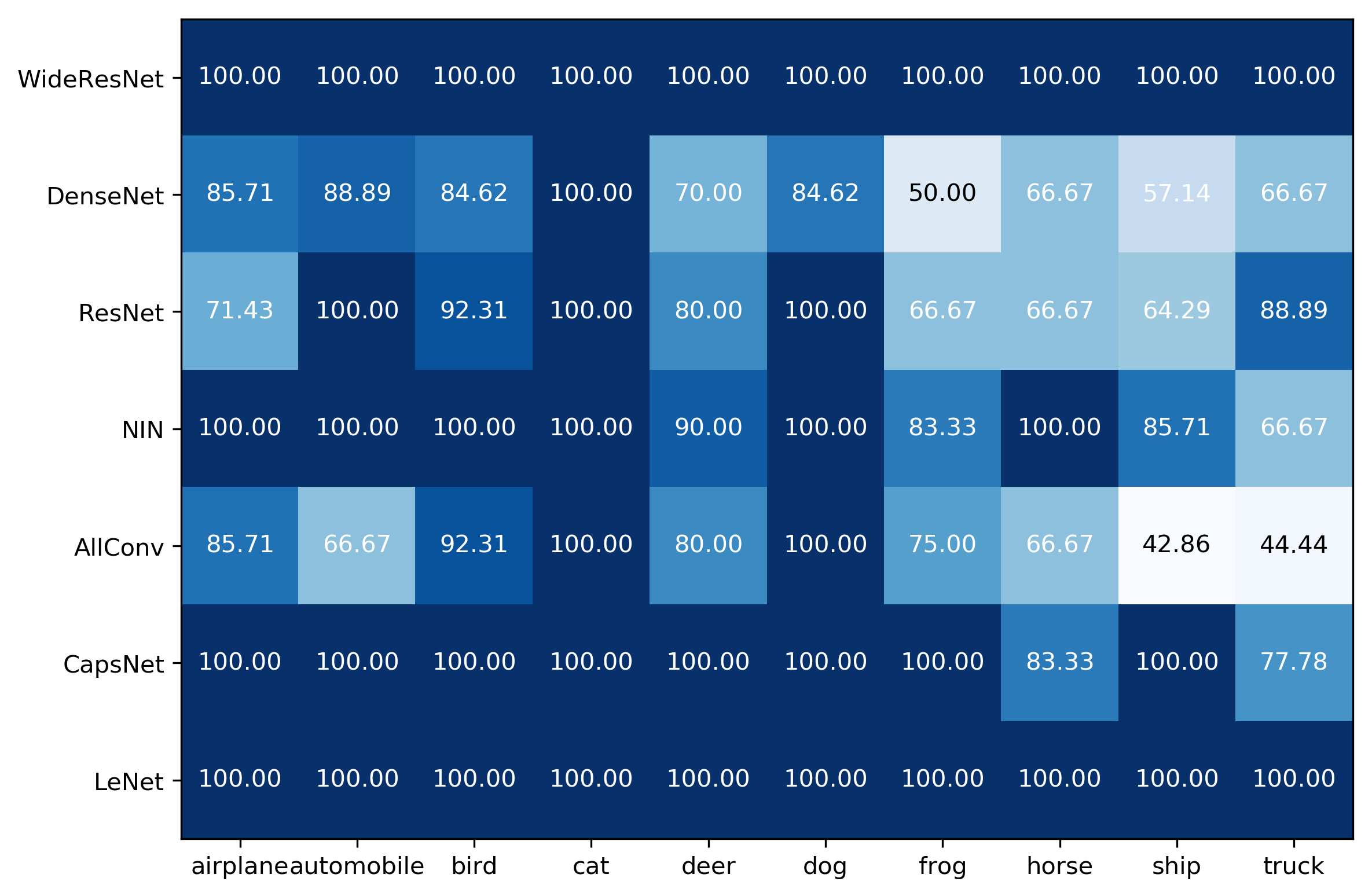}\\
    		\includegraphics[width=0.45\textwidth]{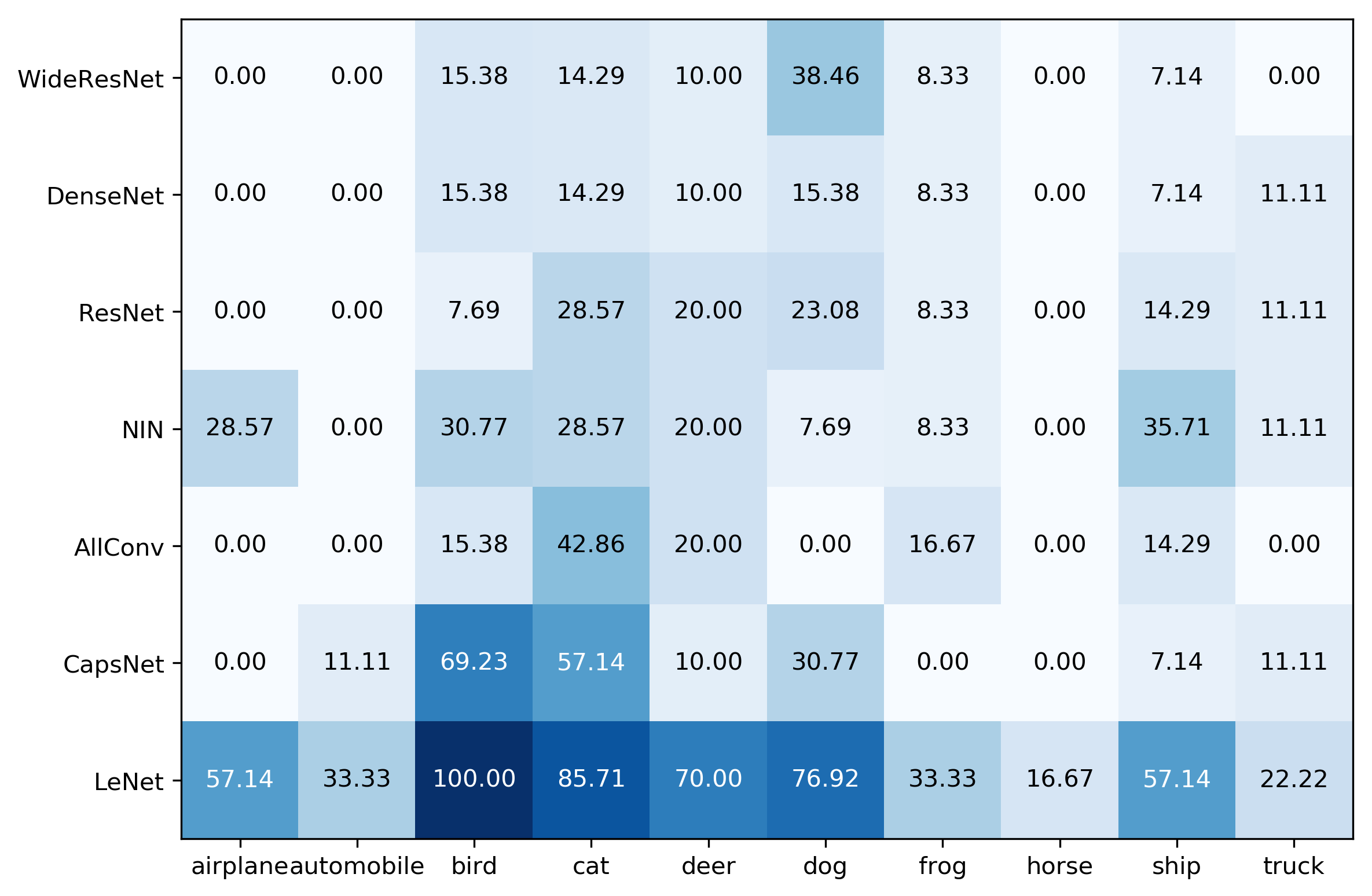}
    		\includegraphics[width=0.45\textwidth]{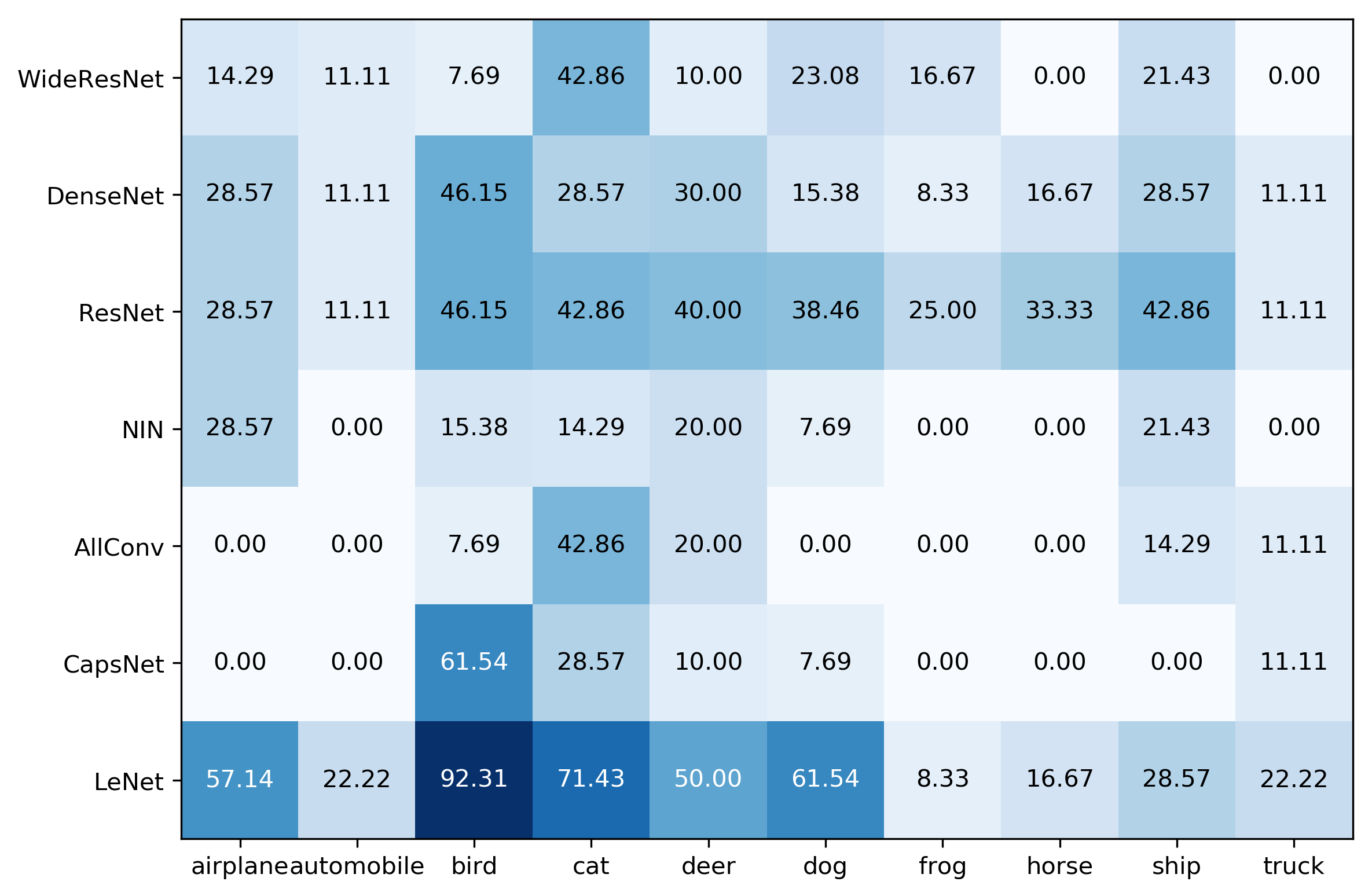}
    		\caption{
    			Adversarial accuracy from Table \ref{table_attack} across classes.
    			The two diagrams at left and right are respectively $L_0$ and $L_\infty$ attacks.
    			The top diagrams used $th=10$ while the bottom ones used $th=1$.
    		}
    		\label{tab_classes}
    	\end{figure*}

        We evaluated our assessed neural networks further against well-known adversarial attacks such as \nocite{art2018}
        Fast Gradient Method (FGM) \cite{goodfellow2014explaining}, 
        Basic Iterative Method (BIM) \cite{kurakin2016adversarial}, 
        Projected Gradient Descent Method (PGD) \cite{madry2017towards}, 
        DeepFool \cite{moosavi2016deepfool}, and 
        NewtonFool \cite{jang2017objective}.
        Please, note that for FGM, BIM, PGD attacks $\epsilon=8 \text{(Default Value)} \approx th=10$ of $L_\infty$ Attack on our robustness scales.
        While DeepFool and NewtonFool do not explicitly control the robustness scale.
        Table \ref{few_table} compares the existing white-box attacks and black-box attacks with our proposed attacks. 
        Notice that, although all the existing attacks are capable of fooling neural networks. 
        We notice some peculiar results, like DeepFool Attack, was less successful against the LeNet, which was most vulnerable to our proposed attacks (Table \ref{table_attack}). 
        Moreover, ResNet and DenseNet had much better robustness for the existing attacks compared to our attacks.
        
        The objective of this article is not to propose better or more effective attacking methods but rather to propose an assessment methodology, and its related duality conjecture (the necessity of evaluating both $L_0$ and $L_\infty$ Attacks).
        However, the proposed Threshold $L_\infty$ Attack in the assessment methodology is more accurate than other attacks while requiring less amount of perturbation.
        The Threshold Attack requires less perturbation than the One-Pixel attack (only circa $12\%$ of the amount of perturbation of the One-Pixel Attack $th=1$) which was already considered one of the most extreme attacks needing less perturbation to fool neural networks. 
        This sets up an even lower threshold to the perturbation, which is inevitable to fool neural networks.

        Notice that, the behaviour of the existing attacks is similar to our Threshold $L_\infty$ Attack (Table \ref{few_table}).
        This suggests that the current evaluations of the neural networks focus on increasing the robustness based on $L_\infty$ Norm. 
        However, our study shows that behaviour of $L_0$ Norm differs from the $L_\infty$ Norm (Table \ref{few_table}), and the robustness for the $L_\infty$ Norm may not be sufficient to study the robustness and vulnerabilities of the neural networks as a whole.
        
    \subsection{Dependency Of Proposed Adversarial Attacks On Classes}
        
        \begin{figure*}[!t]
            \centering
            \includegraphics[width=0.45\textwidth]{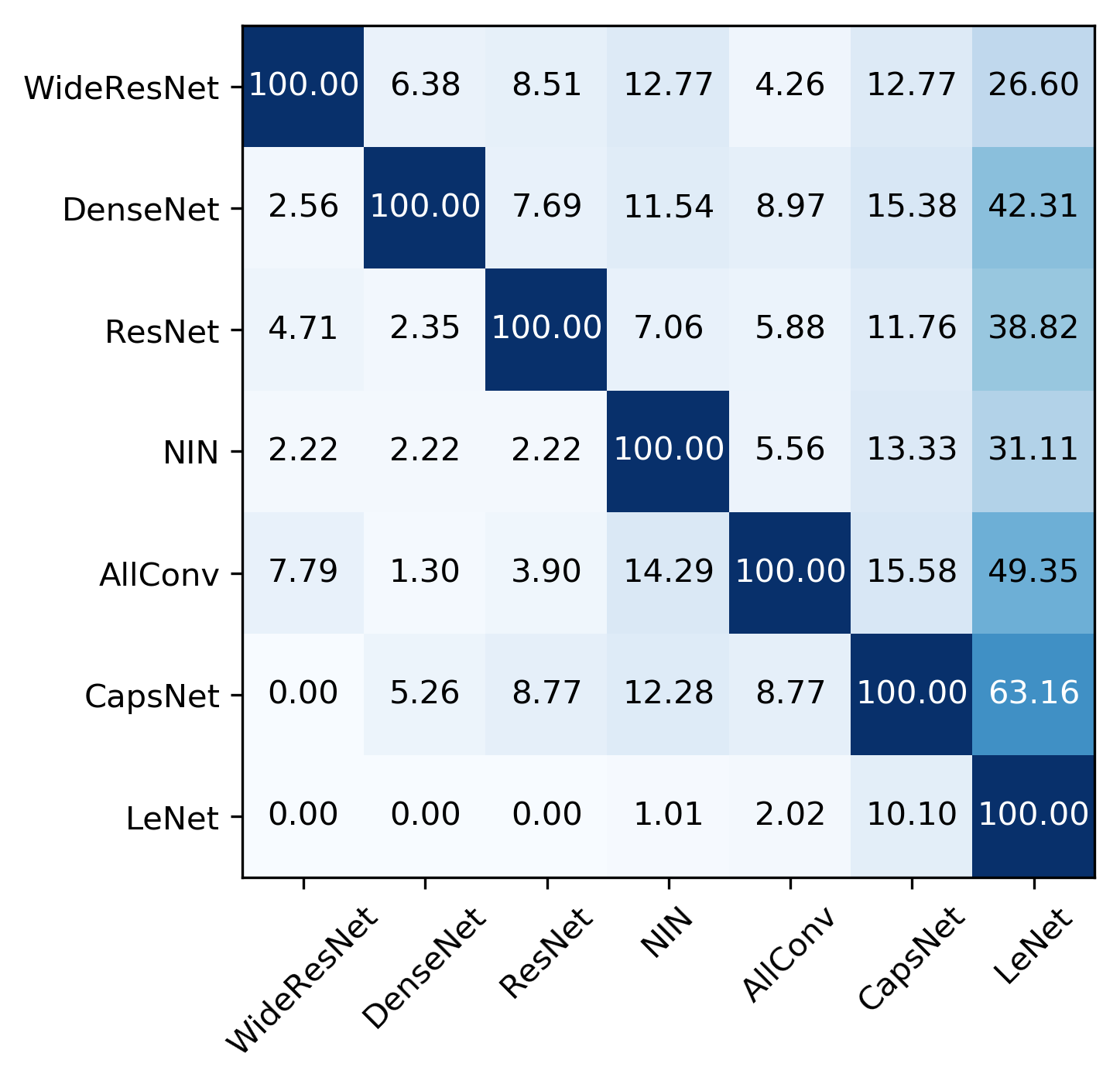}
            \includegraphics[width=0.45\textwidth]{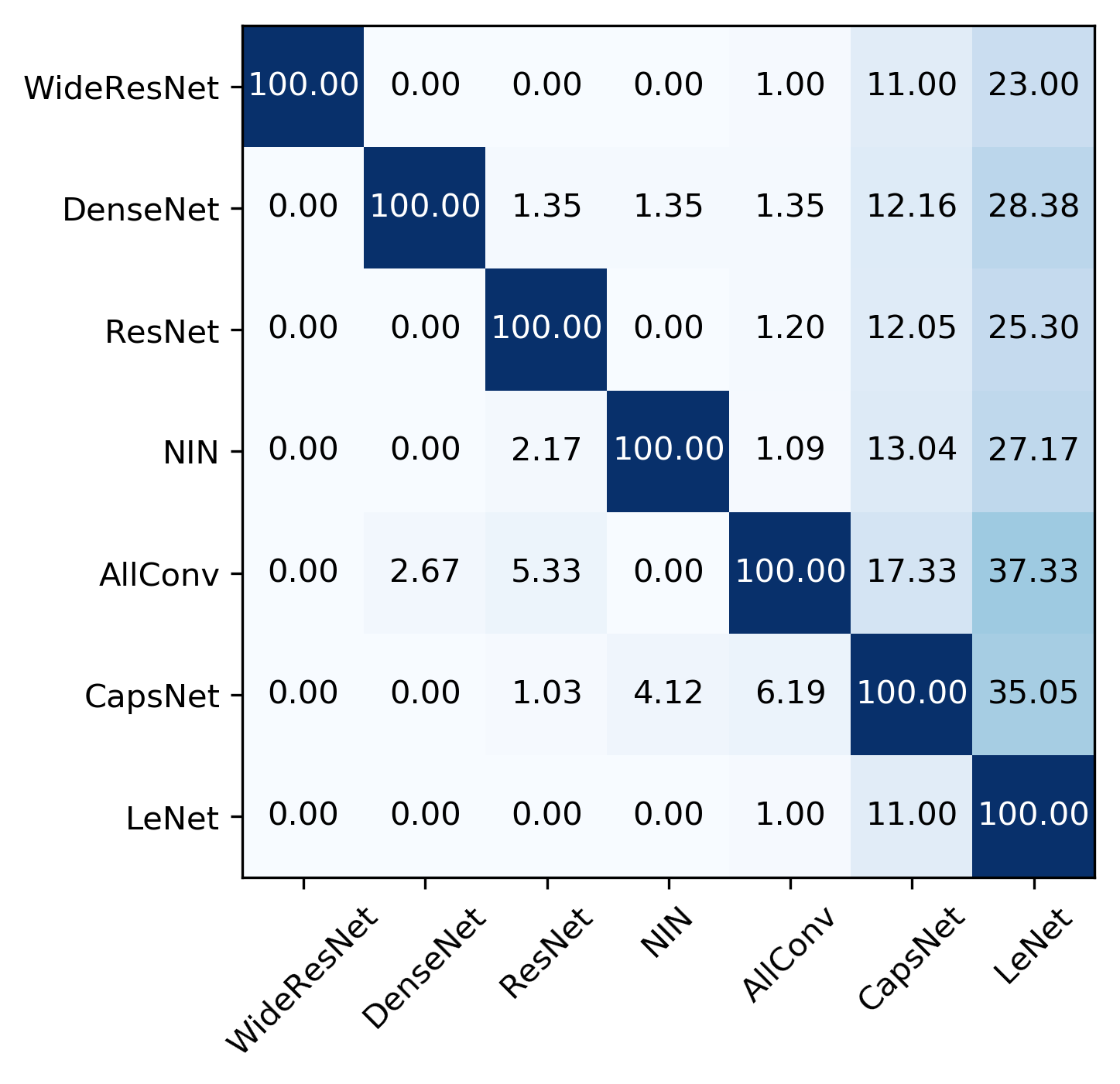}
            \caption{
            Accuracy of adversarial samples when transferring from the a given source model (row) to a target model (column) for both $L_\infty$ black-box Attacks (left) and $L_0$ black-box Attacks (right).
            The source of the adversarial samples is on the y-axis with the target model on the x-axis.
            The adversarial samples were acquired from $100$ original images attacked with $th$ varying mostly from one to ten.
            The maximum value of $th$ is set to $127$. 
            }
            \label{transfer}
        \end{figure*}

        We further separated the adversarial accuracy (Table \ref{table_attack}) into classes (Figure \ref{tab_classes}).
        This is to evaluate the dependence of proposed adversarial attacks on specific classes, 
        Figure \ref{tab_classes} shows an already known feature that some classes are more natural to attack than others.
        For example, the columns for bird and cat classes are visually darker than frog and truck classes for all diagrams.
        This happens because classes with similar features and therefore, closer decision boundaries are more natural to attack.

        Interestingly, the Figure \ref{tab_classes} reveals that neural networks tend to be harder to attack in only a few classes.
        This may suggest that these networks encode some classes far away from others (e.g., projection of the features of these classes into a different vector).
        Consequently, the reason for their relative robustness may lie on a simple construction of the decision boundary with a few distinct and sharply separated classes. 

    \subsection{Extremely Fast Quality Assessment: Transferability of Adversarial Samples}
    \label{section_transferability}

        If adversarial samples from one model can be used to attack different models and defences, it would be possible to create an ultra-fast quality assessment.
        Figure \ref{transfer} shows that indeed, it is possible to qualitatively assess a neural network based on the transferability of adversarial samples.

        Beyond being a faster method, the transferability of samples has the benefit of ignoring any masking of gradients which makes hard to search but not to transfer. 
        This shows that the vulnerability in neural networks is still there but hidden.
        Interestingly, the transferability is mostly independent on the type of attack ($L_0$ or $L_\infty$), with most of the previously discussed differences disappearing.
        There are some differences like $L_0$ attacks are less accurate than most of the $L_\infty$ ones.
        This suggests that positions of pixel and their variance are relatively more model-specific than small changes in the whole image.

        Generally speaking, transferability is a quick assessment method which, when used with many different types of adversarial samples, gives an approximation of the model's robustness.
        This approximation is not better or worse but different.
        It differs from usual attacks because 
        (a) it is not affected by how difficult it is to search adversarial samples, taking into account only their existence, and 
        (b) it measures the accuracy to commonly found adversarial samples rather than all searchable ones.

        Therefore, in the case of low $th$ values, transferability can be used as a qualitative measure of robustness. 
        However, its values are not equivalent to or close to real adversarial accuracy.
        Thus, it serves only as a lower bound.

    \subsection{Adversarial Sample Distribution of Quality Assessment}
    \label{section_distribution}

        \begin{figure}[!t]
            \centering
            \includegraphics[width=0.23\textwidth]{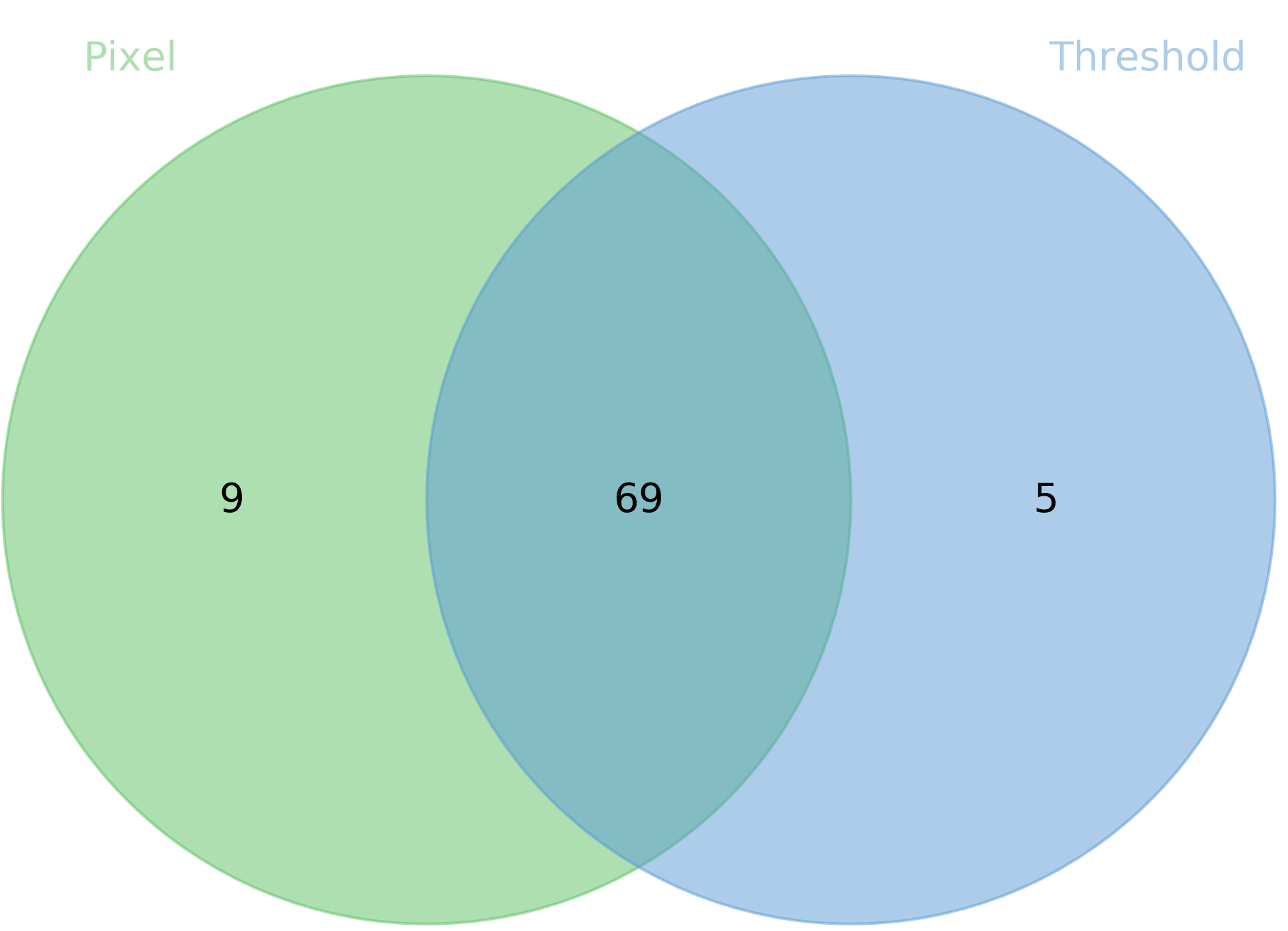}
            \includegraphics[width=0.24\textwidth]{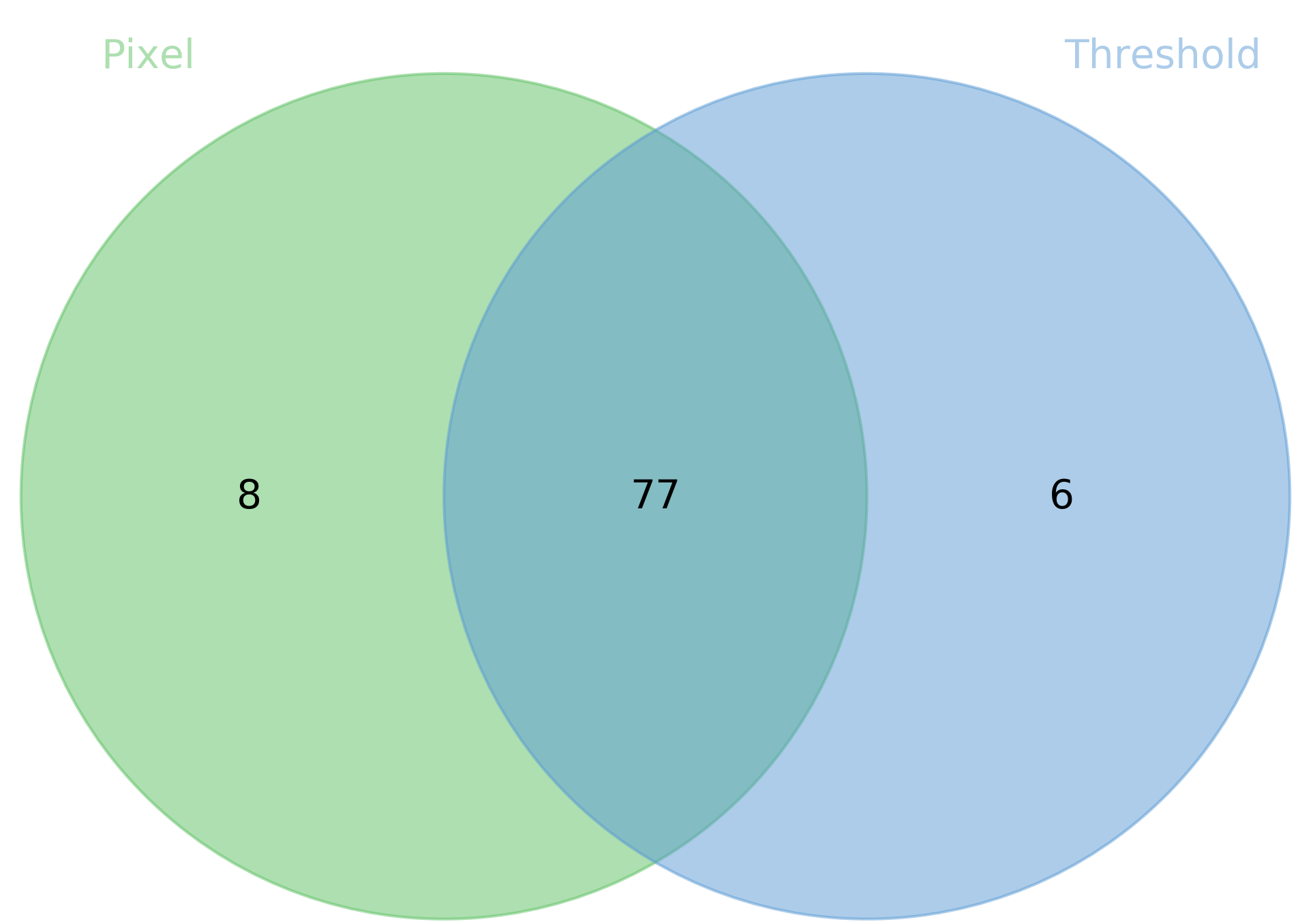}
            \caption{Distribution of adversarial samples found on DenseNet (left) and ResNet (right) using $th=10$ with both few-pixel ($L_0$) and threshold ($L_\infty$) Attacks.}
            \label{dist}
        \end{figure} 
    
    	\begin{figure*}[!t]
    		\centering
    		\includegraphics[width=0.45\textwidth]{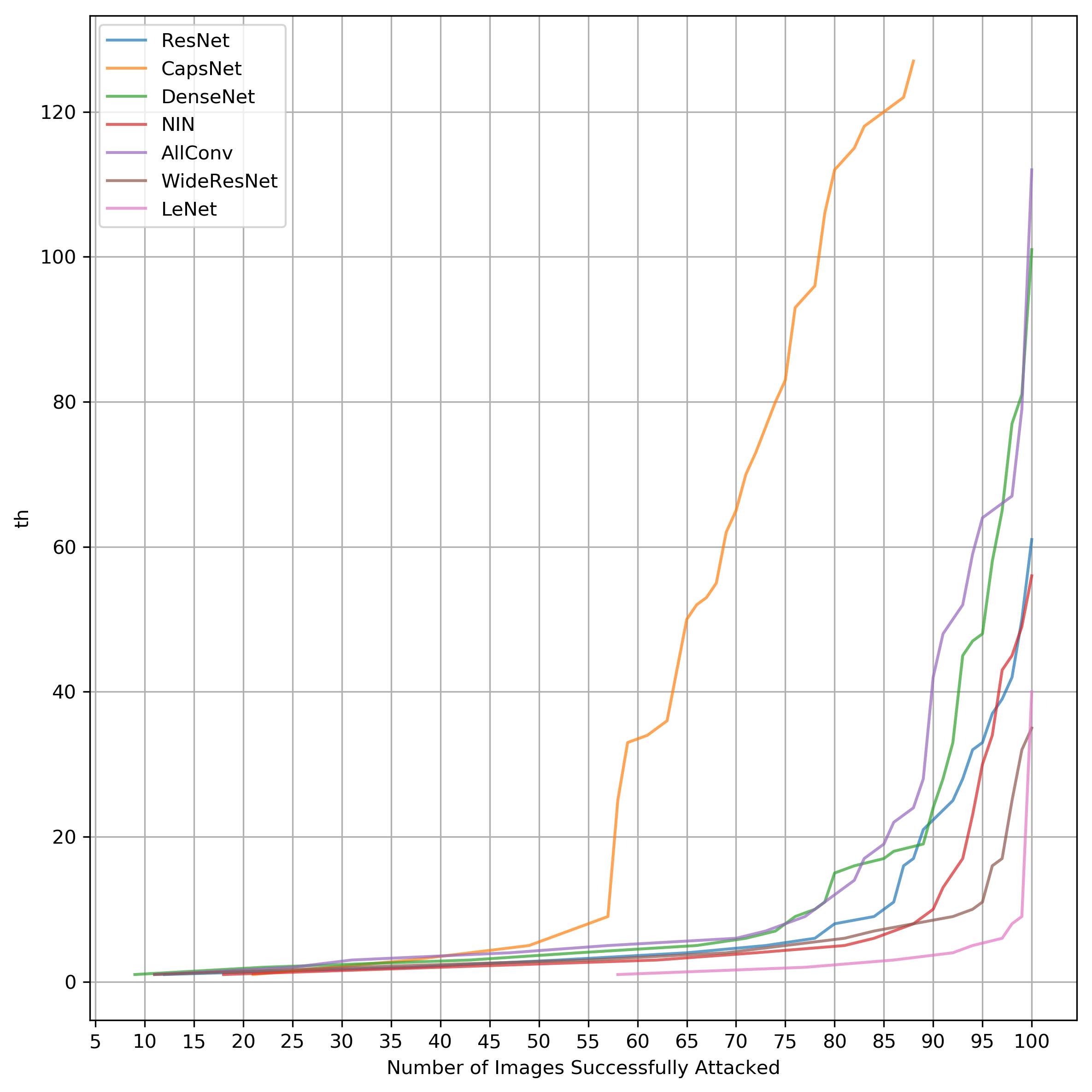}
    		\includegraphics[width=0.45\textwidth]{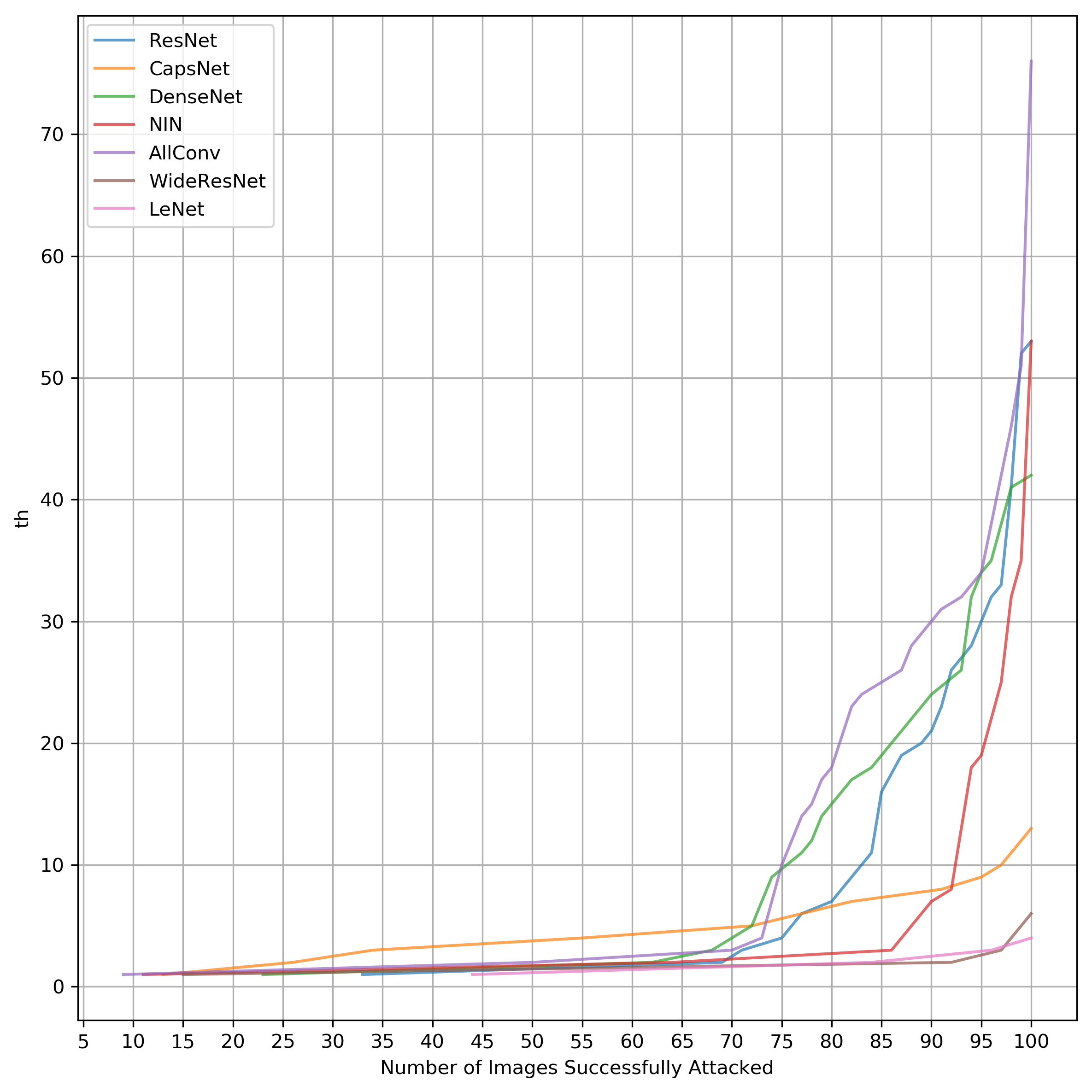}
    		\caption{Adversarial accuracy per $th$ for $L_0$ and $L_\infty$ Attack.}
    		\label{deep}
    	\end{figure*}
    	
    	\begin{table}[!t]
    		\centering
    		\resizebox{0.75\columnwidth}{!}{%
    			\begin{tabular}{l|cc}
    				\toprule
    				\textbf{Model}      & \textbf{$\mathbf{L_0}$ Attack}     & \textbf{$\mathbf{L_\infty}$ Attack}\\
    				\midrule
    				WideResNet &  425.0           &  141.5\\
    				DenseNet   &  989.5           &  696.0\\
    				ResNet     &  674.0           &  575.5\\
    				NIN        &  528.0           &  364.0\\
    				AllConv    &  1123.5          &  \textbf{849.0}\\
    				CapsNet    &  \textbf{2493.0} &  404.5\\
    				LeNet      &  137.5           &  104.0\\
    				\bottomrule
    			\end{tabular}
    		}
    		\caption{Area under the curve (AUC) for both Few-Pixel ($L_0$) and Threshold ($L_\infty$) black-box Attacks}
    		\label{area}
    	\end{table}
    	
        To understand the importance of the duality for the proposed quality assessment.
        We analyse the distribution of our proposed attacks across samples.
        In some cases, the distribution of samples for $L_0$ and $L_\infty$ can be easily verified by the difference in adversarial accuracy. 
        For example, CapsNet is more susceptible to $L_\infty$ than $L_0$ types of attacks  while for adversarial training \cite{madry2017towards} the opposite is true (Table \ref{table_attack}).
        Naturally, adversarial training depends strongly on the adversarial samples used in training,
        Therefore, different robustness could be acquired depending on the type of adversarial samples used.

        Moreover, the distribution shows here that even when adversarial accuracy seems close, the distribution of $L_0$ and $L_\infty$ Attacks may differ.
        For example, the adversarial accuracy on ResNet for both $L_0$ and $L_\infty$ with $th=10$ differ by mere $2\%$.
        However, the distribution of adversarial samples shows that around $17\%$ of the samples can only be attacked by either one of the attack types (Figure \ref{dist}).
        Thus, the evaluation of both $L_0$ and $L_\infty$ are essential to verify the robustness of a given neural network or adversarial defence.
        Moreover, this is true even when a similar adversarial accuracy is observed.

    \subsection{Analysing effect of threshold $th$ on learning systems}
    \label{section_analysing_systems}

        To evaluate how networks behave with the increase in threshold, we plot here the adversarial accuracy with the increase of $th$ (Figure \ref{deep}).
        These plots reveal an even more evident difference of behaviour for the same method when attacked with either $L_0$ or $L_\infty$ norm of attacks.
        It shows that the curve inclination itself is different.
        Therefore, $L_0$ and $L_\infty$ Attacks scale differently.
        
        From Figure \ref{deep}, two classes of curves can be seen.
        CapsNet behaves on a class of its own while the other networks behave similarly.
        CapsNet, which has an entirely different architecture with dynamic routing, shows that a very different robustness behaviour is achieved.
        LeNet is justifiably lower because of its lower accuracy and complexity.
        
        To assess the quality of the algorithms in relation to their curves, the Area Under the Curve (AUC) is calculated by the trapezoidal rule.
         defined as:
        $\text{AUC} = \Delta n_a \left( \frac{th_1}{2} + th_2 + th_3 + \ldots + th_{n-1} + \frac{th_n}{2} \right)$
        where $n_a$ is the number of images attacked and $th_1, th_2 , \ldots  th_n$ are different values of $th$ threshold for a maximum of $n=127$.
        Table \ref{area} shows a quantitative evaluation of Figure \ref{deep} by calculating the Area Under the Curve (AUC).
        
        There is no network which is robust in both attacks.
        CapsNet is the most robust neural network for $L_0$ attacks while AllConv wins while being followed closely by other neural networks for $L_\infty$.
        Although requiring a lot more resources to be drawn, the curves here result in the same conclusion achieved by Table \ref{table_attack}.
        Therefore, the previous results are a good approximation of the behaviour promptly.

\section{Conclusions}

    In this article, we propose a model agnostic dual quality assessment for adversarial machine learning, especially for neural networks.
    By investigating the various state-of-the-art neural networks as well as arguably the contemporary adversarial defences, it was possible to:
    (a) show that robustness to $L_0$ and $L_\infty$ Norm Attacks differ significantly, which is why the duality should be taken into consideration.
    (b) verify that current methods and defences, in general, are vulnerable even for $L_0$ and $L_\infty$ black-box Attacks of low threshold $th$, and 
    (c) validate the dual quality assessment with robustness level as a good and efficient approximation to the full accuracy per threshold curve.
    Interestingly, the evaluation of the proposed method (Threshold Attack) was shown to require surprisingly less amount of perturbation.
    This novel $L_\infty$ black-box Attack based on CMA-ES required only circa $12\%$ of the amount of perturbation used by the One-Pixel Attack while achieving similar accuracy.
    Thus, this article analyses the robustness of neural networks and defences by elucidating the problems as well as proposing solutions to them.
    Hopefully, the proposed dual quality assessment and analysis on current neural networks' robustness will aid the development of more robust neural networks and hybrids alike. 

\section*{Acknowledgments}

    This work was supported by JST, ACT-I Grant Number JP-50243 and JSPS KAKENHI Grant Number JP20241216.
    Additionally, we would like to thank Prof. Junichi Murata for the kind support without which it would not be possible to conduct this research.


%



\ifCLASSOPTIONcaptionsoff
  \newpage
\fi

\bibliographystyle{IEEEtran}
\bibliography{../adversarial_machine_learning}

\end{document}